%% file: root.tex
\documentclass[letterpaper, 10 pt, conference]{ieeeconf}  

\IEEEoverridecommandlockouts                              

\usepackage{colortbl}
\usepackage{amsmath}
\usepackage{xcolor}
\usepackage{xurl}

\definecolor{green}{rgb}{0,0.5,0}
\usepackage{booktabs}
\usepackage{graphicx}
\usepackage{multirow}

\title{\LARGE \bf

Can Real-to-Sim Approaches Capture Dynamic Fabric Behavior for Robotic Fabric Manipulation?

}

\author{Yingdong Ru$^{1}$, Lipeng Zhuang$^{1}$, Zhuo He$^{1}$, Florent P. Audonnet$^{1}$ and  Gerardo Aragon-Caramasa$^{1}$
\thanks{$^{1}$ School of Computing Science, University of Glasgow, G12 8QQ, Scotland, United Kingdom {\tt\small yingdong.ru@glasgow.ac.uk; gerardo.aragoncamarasa@glasgow.ac.uk}}
}

\begin{document}

\maketitle
\thispagestyle{empty}
\pagestyle{empty}

\begin{abstract}


This paper presents a rigorous evaluation of Real-to-Sim parameter estimation approaches for fabric manipulation in robotics. The study systematically assesses three state-of-the-art approaches, namely two differential pipelines and a data-driven approach. We also devise a novel physics-informed neural network approach for physics parameter estimation. These approaches are interfaced with two simulations across multiple Real-to-Sim scenarios (\textit{lifting}, \textit{wind blowing}, and \textit{stretching}) for five different fabric types and evaluated on three unseen scenarios (\textit{folding}, \textit{fling}, and \textit{shaking}). We found that the simulation engines and the choice of Real-to-Sim approaches significantly impact fabric manipulation performance in our evaluation scenarios. Moreover, PINN observes superior performance in quasi-static tasks but shows limitations in dynamic scenarios.
Videos and source code are available at \url{cvas-ug.github.io/real2sim-study}.


\end{abstract}

\input{sections/intro}

\input{sections/lit_review}
\input{sections/task_desc}

\input{sections/methods}
\input{sections/exp_methodology}
\input{sections/results}
\input{sections/conclusions}


\bibliographystyle{IEEEtran}
\bibliography{References.bib}

\end{document}

%% file: sections/intro.tex
\section{INTRODUCTION}\

Robotic manipulation of garments poses significant challenges due to their inherent properties, a high-dimensional configuration space, and rapid, irregular deformations during handling~\cite{duan2022continuous}. Nowadays, researchers lean towards using simulation environments to develop and train policies for fabric manipulation~\cite{lin2021softgym}. However, they rely on default physics parameters of fabrics in the simulation that fail to accurately replicate real fabric behavior, which contributes to the Sim-to-Real gap~\cite{elguea2023review}\cite{matas2018sim}.

In order to minimize the mismatch between real and simulated fabrics' behavior, current Real-to-Sim (\textit{real2sim}) approaches such as differential pipelines (e.g. DiffCLOUD~\cite{sundaresan2022diffcloud} and  DiffCP~\cite{du2021diffpd}) and data-driven approaches (such as~\cite{DBLP:conf/cvpr/RuniaGSS20} and~\cite{duan2022learning}) aim to estimate the physics parameters of fabrics by aligning the behavior of the real and simulated fabric. Despite these advancements, existing \textit{real2sim} approaches are often validated using the same training scenario. Similarly, optimizing a large number of parameters, particularly those with limited physical significance, can lead to overfitting within a single scenario. Additionally, the absence of ground-truth physics parameters for real fabrics poses a fundamental challenge in evaluating the estimation accuracy~\cite{bouman2013estimating}. 
This raises the question of \textit{whether these approaches can estimate physically meaningful parameters that accurately represent real-world fabric properties and generalize to unseen fabric manipulation tasks in simulation.} 

Physics-Informed Neural Networks (PINNs) have been proposed as an alternative approach that incorporates PDE equations directly into the training process~\cite{DBLP:journals/jscic/CuomoCGRRP22}. While PINNs have been used for parameter estimation of metal plates~\cite{zhou2023transfer}, no application has been reported for fabric parameter estimation to the best of our knowledge. The potential of PINNs for fabric \textit{real2sim} parameter estimation remains unexplored; we, therefore, pose the question of \textit{whether incorporating physics-informed learning can better estimate physics parameters of fabrics.}

\begin{figure}[t]
   \centering
   \includegraphics[width=0.85\linewidth]{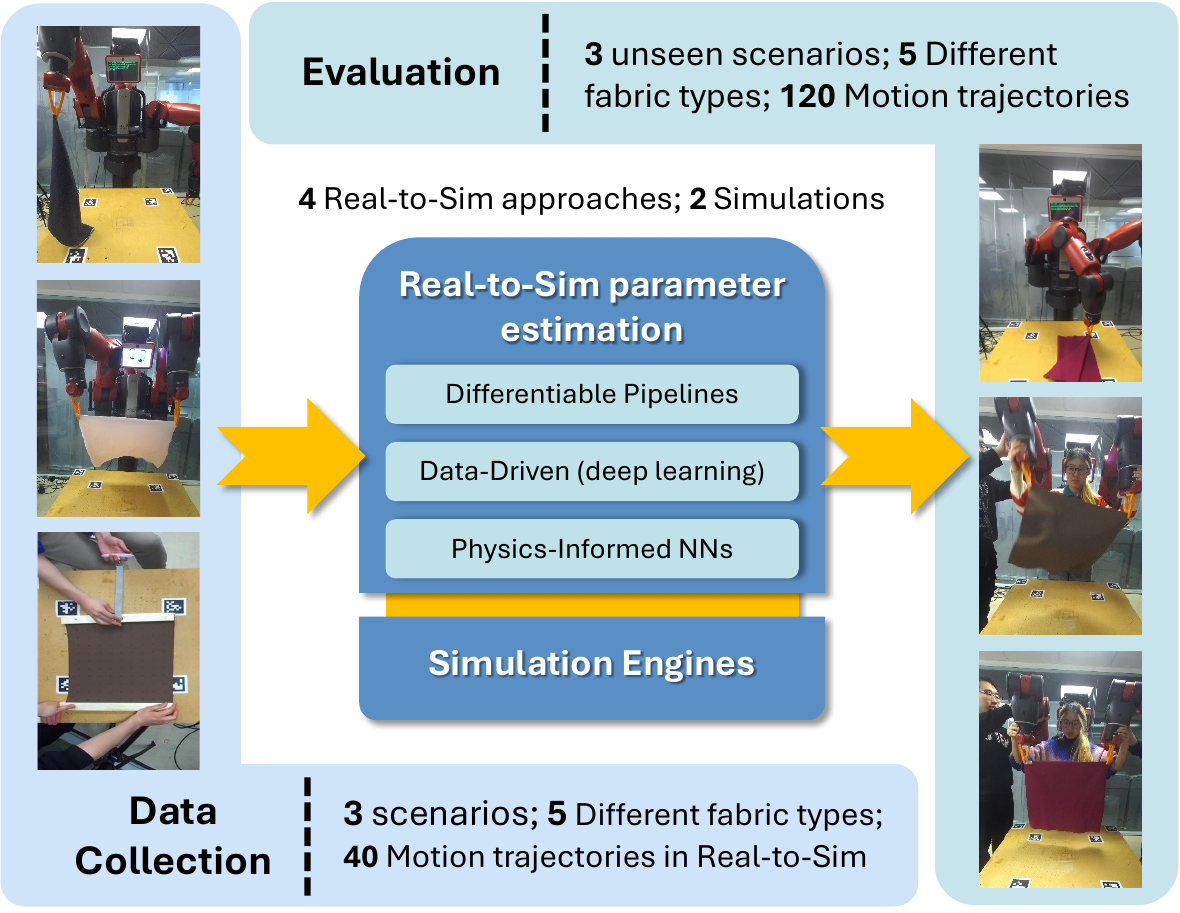}
   \caption{This study focuses on \textit{real2sim} physics parameter estimation for
fabrics and evaluation across different scenarios. First,real-world fabric data is collected from three scenarios with five fabric types and 40 motion trajectories for \textit{real2sim}. Physics parameters are then estimated using differentiable pipelines, deep learning, and physics-informed neural networks (PINNs) and applied within the same simulation engine. Evaluation is performed across three unseen scenarios with 120 motion trajectories, comparing simulated and real-world fabric behavior to assess accuracy and generalization. The study examines the impact of four \textit{real2sim} approaches, two simulation engines, and three \textit{real2sim} scenarios on estimation performance.}\label{fig:first_fig}
\end{figure}

To investigate this further, several factors may influence the reliability of \textit{real2sim} performance. First, different \textit{real2sim} approaches employ distinct optimization strategies, which leads to potential variations in estimation. Second, the effectiveness of these approaches may also heavily depend on the choice of scenarios. Third, the performance of \textit{real2sim} approaches is influenced by the choice of the simulation framework, as each simulator employs its own dynamics engine and equations to describe fabric physics behavior. This raises another research question: \textit{how do different factors—such as \textbf{real2sim} approaches, scenario selection, and simulator choice—affect \textbf{real2sim} performance across diverse scenarios?}



\begin{figure*}[t]
   \centering
   \includegraphics[width=0.9\linewidth]{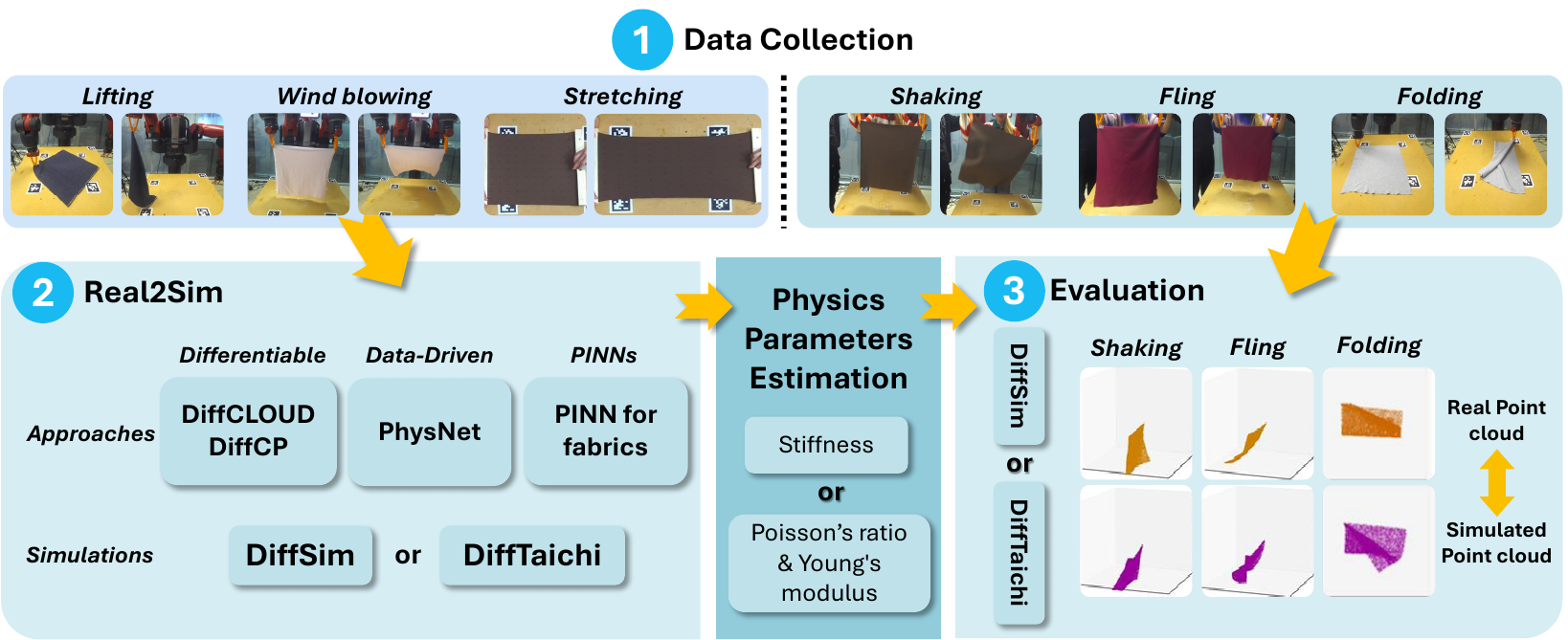}
   \caption{The proposed pipeline consists of three stages: (1) Data Collection, (2) \textit{real2sim} Physics Parameter Estimation of fabrics, and (3) Evaluation. In (1), real-world fabric behavior is recorded across six scenarios—three for \textit{real2sim} (\textit{lifting}, \textit{wind blowing}, \textit{stretching}) and three for evaluation (\textit{shaking}, \textit{fling}, \textit{folding}). In (2), physics parameters are estimated using three approaches: differential pipelines (DiffCLOUD, DiffCP), deep learning (PhysNet~\cite{duan2022learning}), and physics-informed neural networks. Specifically, DiffSim~\cite{DBLP:conf/icml/QiaoLKL20} is used to estimate~\textit{stiffness} and is then employed to simulate fabric behavior using the estimated values, while DiffTaichi~\cite{DBLP:conf/iclr/HuALSCRD20} follows the same process for \textit{Young’s modulus} and \textit{Poisson’s ratio}. Finally, the evaluation stage assesses the generalization of estimated parameters by comparing simulated fabric behaviors to real-world observations across unseen scenarios.\label{fig:main_fig}}
\end{figure*}

To answer these research questions, we conduct 
a rigorous study (as illustrated in Fig.\ref{fig:first_fig}) aimed at evaluating the generalization of three state-of-the-art \textit{real2sim} parameter estimation approaches,
and a novel PINN-based approach for garment physics parameter estimation. 
For this, we deployed these approaches to two simulators, DiffSim~\cite{DBLP:conf/icml/QiaoLKL20} and DiffTaichi~\cite{DBLP:conf/iclr/HuALSCRD20} (Fig. \ref{fig:main_fig}) where we train on three \textit{real2sim} scenarios and evaluate on three unseen, different scenarios using 
single- and dual-arm configurations and five different fabric types.

Our study consists of two stages. In the \textit{real2sim} stage, our goal is to align the physical behavior of simulated fabric with real fabric in order to estimate the fabric's physics parameters. 
In the evaluation stage, we leverage the physics parameters obtained in the \textit{real2sim} stage and then compare the simulated fabric behavior with real-world fabric scenarios, quantifying the gap to evaluate the generalization ability of the \textit{real2sim} approaches. Notably, we are the first to leverage PINNs to solve the \textit{real2sim} problem of physics parameter estimation of fabrics by deriving an anisotropic constitutive model based on the Elastodynamic constitutive law. 

Hence, our contributions are:

\begin{itemize}

\item A rigorous evaluation of the state-of-the-art \textit{real2sim} approaches to study their generalization ability on unseen scenarios.

\item An analysis of the relative impact of various factors on the generalizability of \textit{real2sim} performance across diverse robotic manipulation tasks of fabrics.

\item A new \textit{real2sim} approach based on PINNs to estimate fabric physics parameters.

\item A dataset for fabric manipulation, which includes synchronized RGB-D images, point clouds, and data of six scenarios and five types of fabrics.


\end{itemize}

%% file: sections/lit_review.tex
\section{Literature review}\label{sec:LR}

Differentiable physics simulation consists of finding optimized solutions to inverse problems, including control optimization and motion planning. These simulations can calculate gradients in relation to various input parameters~\cite{hu2019chainqueen}. Moreover, it can optimize the physics parameters of materials to align to a target trajectory of objects~\cite{arnavaz2023differentiable}. Differentiable physics simulation eliminates the need for pre-training and specialized equipment, adheres to the fundamental principles of physics, and enables faster convergence in learning-based tasks~\cite{arnavaz2023differentiable}.

While there exist differentiable simulations supporting soft bodies and elastoplastic materials such as Gradsim ~\cite{DBLP:conf/iclr/MurthyMGVPWCPXE21}, Chainqueen~\cite{hu2019chainqueen} and Diffpd~\cite{du2021diffpd}, differentiable simulations specifically designed for fabric simulation are scarce. Only three are currently available: DiffCloth~\cite{li2022diffcloth}, DiffSim~\cite{DBLP:conf/icml/QiaoLKL20} and DiffTaichi~\cite{DBLP:conf/iclr/HuALSCRD20}. Specifically, DIFFCLOUD~\cite{DBLP:journals/corr/abs-2007-08501} differential pipeline leverages DiffSim to simulate fabrics via a Finite Element Method (FEM) coupled with a differentiable renderer. While the DiffCP differential pipeline~\cite{diffcp2019} employs DiffTaichi as a differentiable simulation which is based on the Material Point Method (MPM)~\cite{DBLP:conf/iclr/HuALSCRD20}. In this paper, we implement DIFFCLOUD and DiffCP as the differential pipelines. However, we exclude all position-based dynamics simulators, such as DiffCloth~\cite{li2022diffcloth}, since these simulators prioritize numerical stability and real-time performance by directly manipulating positions to satisfy constraints that compromise physical accuracy in terms of energy and momentum conservation.

In addition to differentiable physics simulations, data-driven approaches for parameter estimation of fabrics have been proposed in the literature. For instance, 
Runia \textit{et al.}~\cite{DBLP:conf/cvpr/RuniaGSS20} developed an approach to estimate the physical properties of real objects by computing a physics similarity distance via a pairwise contrastive loss between simulated and real fabrics. Then, their approach minimizes this distance by adjusting simulation parameters via Bayesian optimization.
We selected PhysNet~\cite{duan2022learning} as our deep learning baseline for \textit{real2sim} estimation because it improves on the framework of~\cite{DBLP:conf/cvpr/RuniaGSS20} by integrating depth information to better capture the dynamic behavior of fabrics and garments, and employs a triplet contrastive loss for enhanced feature discrimination.


In the literature, there have been approaches that integrate physics laws as a regularizer in neural network training, termed Physics-Informed Neural Networks (PINNs). That is, Partial Differential Equations (PDEs) are coupled into the neural network's learning process~\cite{DBLP:journals/corr/abs-2402-00326}. PINNs optimize both data-driven and physics-based losses derived from governing physics equations' residuals, enabling physically consistent predictions and improved generalization with limited or noisy data~\cite{DBLP:journals/jcphy/ZhangLL25}.
While PINNs have been widely used to solve the Navier-Stokes PDEs for predicting the pressure and velocity in fluid dynamics~\cite{DBLP:journals/corr/abs-2304-03689}, Xu et al.~\cite{DBLP:conf/iclr/0028MNRMGM22} have extended this methodology to elastodynamics PDEs, facilitating parameter estimation for elastic-plastic materials, such as metal plates and beams. They demonstrated that this approach is particularly useful in scenarios where data acquisition is limited, lacks ground truth, and is costly to test these materials. In this paper, we build on~\cite{DBLP:conf/iclr/0028MNRMGM22} to devise a \textit{real2sim} approach for fabrics by using a constitutive law that defines the relationship between stress and strain and describes how a fabric deforms under applied forces.

%% file: sections/task_desc.tex
\section{Task description}\label{sec:task_desc}

For \textit{real2sim}, three scenarios—\textit{lifting}, \textit{stretching}, and \textit{wind blowing}—were selected, mirroring those adopted in previous \textit{real2sim} studies to provide a standardized framework for comparison. In the \textit{lifting} scenario, a robotic gripper grabs the left corner of a fabric that is initially flat on a table and lifts it. In the \textit{stretching} scenario, the fabric's two sides are clamped with wooden strips; one side remains fixed, while the other is connected to a tensile device that pulls it to a predetermined distance before holding it to measure the force. In the \textit{wind blowing} scenario, the top two corners of the fabric are grasped, and a fan positioned in front of the fabric applies a wind force to its central region, with wind speeds measured by an electronic anemometer. 

For evaluation, three additional scenarios were designed {folding}, \textit{fling}, and \textit{shaking}. In the \textit{folding} scenario, a random corner of the fabric is folded towards its opposite side. In the \textit{fling} scenario, two robotic grippers first grasp the fabric from the top corners and then undergo rapid acceleration, causing the fabric to dynamically fling and flap. In the \textit{shaking} scenario, two robotic grippers again grasp the fabric at the top corners. One gripper then oscillates vertically, causing the fabric to shake dynamically. 

%% file: sections/methods.tex
\section{Methods}\label{sec:methods}

The \textit{real2sim} and evaluation scenarios in Sec \ref{sec:task_desc} enable us to establish a framework for evaluating the generalization of \textit{real2sim} approaches. This section thus presents the differentiable simulation pipelines and data-driven approaches benchmarked against these scenarios, alongside a novel \textit{real2sim} approach based on the governing Partial Differential Equations (PDEs) for deformable objects using two different simulations. In our experiments, we use five fabric materials with different physics properties.



\subsection{Differential pipelines}
\subsubsection{DIFFCLOUD~\cite{sundaresan2022diffcloud}}\label{sec:Diffcloud1} proposes an optimization framework that aligns the behavior of simulated fabrics with real fabrics by tuning physics parameters (i.e. stiffness parameters). By using a differentiable simulation (DiffSim~\cite{DBLP:conf/icml/QiaoLKL20}) to reconstruct the dynamics of real fabrics in the simulated environment, DIFFCLOUD backpropagates a spatial loss for the optimization of the stiffness parameters. That is, they use a FEM method to capture small deformations for structurally stable scenarios to propagate gradients through mesh nodes. The simulation models the fabric using an anisotropic material framework, where different stiffness parameters characterize the material's behavior along different directional axes. A one-sided-chamfer distance loss is calculated between the randomly sampled simulated point clouds and real point clouds from fabrics to measure the distance between them to enable end-to-end optimization from real point cloud observations to simulation parameters~\cite{sundaresan2022diffcloud}.


\subsubsection{DiffCP~\cite{diffcp2019}}\label{sec:diffcp} employs an optimization pipeline using the DiffTaichi simulator~\cite{DBLP:conf/iclr/HuALSCRD20} to compare simulated point clouds to real observations using one-sided chamfer distance loss; similar to Sec~\ref{sec:Diffcloud1}. DiffTaichi excels at handling extreme deformations, including fractures and fragmentation, via a Material Point Method (MPM). The system optimizes material parameters using Young's modulus and Poisson's ratio via automatic differentiation and backpropagation.


\subsection{Data-Driven - PhysNet~\cite{duan2022learning}}\label{sec:physnet}
we choose PhysNet (as discussed in Sec \ref{sec:LR}) as the data-driven approach in this paper. Specifically, PhysNet is a Siamese network that clusters fabric data based on physical properties. The network uses a feature extraction backbone to extract features to map input samples into a Physics Similarity Map (PSM) in order to measure their physics similarity. PhysNet learns to distinguish fabric samples by their physical characteristics by using a triplet loss function, where each input consists of three images: an anchor, a positive sample (similar physics properties of fabrics), and a negative sample (different physics properties of fabrics). 
Then, Bayesian optimization is used to minimize the physics similarity between simulated and real fabric behaviors. 

\subsection{Physics-Informed Neural Networks (PINNs)}\label{sec:pinn}

Inspired by the Zhou et al.~\cite{zhou2023transfer} (as discussed in Sec \ref{sec:LR}, our objective is to leverage PINNs for \textit{real2sim} parameter estimation by solving the governing PDE/ODE equations solved by simulators. For this, we devised two PINN frameworks for DiffSim and DiffTaichi as follows.

\subsubsection{PINN using DiffSim~\cite{DBLP:conf/icml/QiaoLKL20}\label{sec:pinn-diffsim}} We adapted DiffSim's elastodynamic PDE, which captures fabric behavior than the simpler PDE used for metal plates. The Elastodynamic PDE~\cite{DBLP:journals/corr/abs-2006-08472} equation is:
\begin{equation}
\rho_0 \ddot{\phi} = \nabla \cdot \mathbf{P} + \rho_0 \mathbf{b}
\label{eq:Elastoplastic}
\end{equation}
where, the deformation map $\phi$ describes a vector field, mapping its initial position $X$ to its current position $x$ at time $t$. The density is represented by $\rho_0$ and $b$ is force. $\ddot{\phi}$ is acceleration and $P$ is stress. To be more specific, the stress in the fabric is characterized along three axes, corresponding to the product of the stiffness parameter and the strain~\cite{wang2011data}, defined as:
\begin{equation}
\begin{aligned}
\sigma_{xx} &= C_{11} \cdot \epsilon_x + C_{12} \cdot \epsilon_y, \\
\sigma_{xy} &= C_{33} \cdot \epsilon_{xy}, \\
\sigma_{yy} &= C_{12} \cdot \epsilon_x + C_{22} \cdot \epsilon_y
\end{aligned}
\label{eq:oneLabel}
\end{equation}
$C_{11}$, $C_{12}$, $C_{22}$ and $C_{33}$ represent four stretching stiffness parameters in the weft, diagonal, wrap, and shear directions, respectively. $\epsilon_x$, $\epsilon_y$ and $\epsilon_{xy}$ represent different strains in each direction.

Hence, the input to the PINN consists of the initial positions of nodes ($x$ and $y$) and time ($t$), while the output represents deformation in the $x$ and $y$ directions. The data loss is defined as the MSE between the predictions and the ground truth. 
The predicted output of the neural network, denoted as $\phi$, is differentiated twice with respect to time to obtain acceleration as in Eq. \ref{eq:pinn_group1}, and differentiated with respect to the initial location to obtain strain as in Eq. \ref{eq:pinn_group2}.
\begin{equation}
\begin{array}{cccc}
\displaystyle \frac{\partial \phi_x}{\partial t} = V_x, & 
\displaystyle \frac{\partial \phi_y}{\partial t} = V_y, & 
\displaystyle \frac{\partial v_x}{\partial t} = a_x, & 
\displaystyle \frac{\partial v_y}{\partial t} = a_y
\end{array}
\label{eq:pinn_group1}
\end{equation}
\begin{equation}
\begin{array}{ccc}
\displaystyle \frac{\partial \phi_x}{\partial x} = \epsilon_x, & 
\displaystyle \frac{\partial \phi_y}{\partial y} = \epsilon_y, & 
\displaystyle 0.5\Bigl(\frac{\partial \phi_x}{\partial y} + \frac{\partial \phi_y}{\partial x}\Bigr) = \epsilon_{xy}
\end{array}
\label{eq:pinn_group2}
\end{equation}

The parameters $C_{11}$, $C_{12}$, $C_{22}$ and $C_{33}$ are treated as trainable parameters, and the divergence of the stress ($\nabla$) is computed as:
\begin{equation}
\begin{array}{ccc}
\nabla_{\text{stress}, x} = \frac{\partial \sigma_{xx}}{\partial x} + \frac{\partial \sigma_{xy}}{\partial y}, &\quad &
\nabla_{\text{stress}, y} = \frac{\partial \sigma_{yy}}{\partial y} + \frac{\partial \sigma_{xy}}{\partial x}.
\end{array}
\label{eq:pinn_group3}
\end{equation}

Then the PDE loss can be defined as:
\begin{equation}\label{eq:pinn_group4}
\begin{aligned}
L_{PDE,x} &= \rho_0 \cdot a_x - \left(\frac{\partial \sigma_{xx}}{\partial x} + \frac{\partial \sigma_{xy}}{\partial y}\right) - b\\[1ex]
L_{PDE,y} &= \rho_0 \cdot a_y - \left(\frac{\partial \sigma_{yy}}{\partial y} + \frac{\partial \sigma_{xy}}{\partial x}\right) - b
\end{aligned}
\end{equation}

The objective is to minimize both the PDE loss and the data loss simultaneously. By doing so, we can converge on estimates of the stiffness parameters $C_{11}$, $C_{12}$, $C_{22}$, and $C_{33}$ (trainable parameters).

\subsubsection{PINN for DiffTaichi}\label{sec:pinn-difftaichi}

The difference between DiffSim and DiffTaichi lies in their constitutive model definition for fabrics. That is, DiffTaichi uses MPM to solve elastoplastic PDEs, where the governing PDE equations are:

\begin{equation}
\begin{aligned}
\sigma_{xx} &= \frac{E}{(1+\nu)(1-2\nu)}\Big[(1-\nu)\epsilon_x + \nu\epsilon_y\Big],\\[1mm]
\sigma_{yy} &= \frac{E}{(1+\nu)(1-2\nu)}\Big[\nu\epsilon_x + (1-\nu)\epsilon_y\Big],\\[1mm]
\sigma_{xy} &= \frac{E}{2(1+\nu)}\epsilon_{xy}.
\end{aligned}
\label{eq:sigma}
\end{equation}
where $E$ is the Young's modulus and $\nu$, Poisson's ratio. Moreover, DiffTaichi adopts an isotropic material model, meaning that the physical behavior of the fabric remains consistent in all directions, as shown in Eq~\ref{eq:sigma}. In this model, $E$ and $\nu$ are treated as trainable parameters within the PDE loss function, and we use the same framework as described in Sec.~\ref{sec:pinn-diffsim}.

%% file: sections/exp_methodology.tex
\section{Experimental Methodology}


\subsection{Scenario Settings}\label{sec:pinn2}



We use the \textit{real2sim} \textit{lifting} scenario for DIFFCLOUD, DiffCP, and PhysNet while estimating the physics parameters of the fabrics. However, for PINNs, the \textit{lifting} scenario presents significant challenges because they require the tracking of corresponding points from the initial to the final frame, which is a challenging task while tracking stable points due to shape alterations in all directions--we attempted to use SOTA dense point tracking approaches but failed at consistently tracking points~\cite{DBLP:conf/iccv/DoerschYVG0ACZ23}. Similarly, we use the \textit{wind blowing} scenario in DIFFCLOUD and PhysNet. DiffCP is not applicable in the \textit{wind blowing} scenario, as its simulator does not support external wind forces. Furthermore, the non-uniform force distribution caused by wind complicates the body force component of the PDE loss within the PINN framework.

\begin{table}[t]
\centering
\caption{Overview of the Real2Sim training and Evaluation setup}
\label{tab:Overview}
\resizebox{0.45\textwidth}{!}{%
\begin{tabular}{@{}lccccc@{}}
\toprule
\multicolumn{4}{c}{\textbf{\textit{real2sim} (train)}} & \multicolumn{1}{c}{\textbf{\textit{Evaluation} (test)}} \\
\cmidrule(r){1-4}\cmidrule(r){5-5}

Categories & Approaches & Simulators & Scenarios & Scenarios \\
\midrule
\multirow{2}{*}{Differential pipelines} 
  & DIFFCLOUD & DiffSim    & Lifting, Wind blowing & Folding, Shaking, Fling \\ 
  & DiffCP    & DiffTaichi & Lifting          & Folding, Shaking, Fling \\ \midrule
\multirow{2}{*}{Deep learning} 
  & PhysNet   & DiffSim & Lifting, Wind blowing & Folding, Shaking, Fling \\ 
  & PhysNet       & DiffTaichi        & Lifting           &  Folding, Shaking, Fling\\ \midrule
\multirow{2}{*}{PINN} 
  & PINN for fabrics & DiffSim & Streching & Folding, Shaking, Fling \\
  & PINN for fabrics      & DiffTaichi       & Stretching           & Folding, Shaking, Fling \\
\bottomrule
\end{tabular}%
}
\end{table}

We only use the \textit{stretching} scenario for PINNs because this scenario approximates a quasi-homogeneous force field applied to the fabric, which simplifies the formulation of the body force term in Eq. \ref{eq:Elastoplastic} and yields a predictable deformation pattern that facilitates reliable point tracking, ultimately reducing the inverse problem's complexity and improving convergence. However, differential pipelines and PhysNet do not support \textit{stretching} scenarios.

To evaluate the generalization of fabric behavior, we conducted experiments involving \textit{folding}, \textit{shaking}, and \textit{fling} scenarios (evaluation scenarios). These experiments compared the physical behavior of real fabrics with that of simulated fabrics trained using a \textit{real2sim} scenarios. The evaluation was performed by following identical motion trajectories across five distinct fabric types as described in Sec. \ref{sec:data-collection}. As shown in the Table \ref{tab:Overview}, DIFFCLOUD (Sec. \ref{sec:Diffcloud}) uses DiffSim as the backend simulator, while DiffCP (Sec. \ref{sec:Diffcpsix}) uses DiffTaichi. To ensure consistency and facilitate fair comparison, PhysNet (Sec. \ref{sec:physnet}) and PINNs (Secs. \ref{sec:pinn-diffsim} \& \ref{sec:pinn-difftaichi}) use both DiffSim and DiffTaichi for training and evaluation. 


\subsection{Data collection}\label{sec:data-collection}

We collected data using five fabrics of identical size (45 cm x 45 cm) but with varying physical properties, ranging from highly deformable to shape-retaining. The fabrics used were \textit{camel ponte roma}, \textit{black denim}, \textit{grey interlock}, \textit{pink solid} and \textit{red jet set}; commonly available in textile shops. 
Our hardware setup mirrored that of~\cite{DBLP:flatnfold}. That is, two ZED2i cameras positioned at the front and top to capture RGB-D image sequences at 15 Hz. Specifically for the \textit{stretching} scenario, we positioned an additional ZED2 camera right above the fabric recording at 30 Hz in order to mitigate inaccuracies in point measurement caused by camera tilting. We used SAM2~\cite{DBLP:journals/corr/abs-2408-00714} for fabric segmentation, ensuring that only the observed fabric was retained. Point cloud sequences were then generated from synchronized RGB and depth images based on the camera's intrinsic and extrinsic parameters. For the action information, we used a Rethink Robotics Baxter robot~\cite{trimble2019slip}. For \textit{shaking} and \textit{fling} scenarios, which require high speed, we leveraged Baxter’s zero-G mode~\cite{armcontrolsystem}. For the other scenarios, we used predefined trajectories with linear interpolation to maintain constant speed. The position and rotation of the grippers were tracked at 15 Hz. Overall, our dataset contains synchronized RGB-D images, point clouds, and action information, which is around 50 GB.



\subsection{Performance metrics for Sim-to-Real}

After \textit{real2sim} parameter estimation of fabrics, the reality gap is measured by comparing real point cloud sequences and simulated point cloud sequences by using the same trajectory in the simulations. We chose Chamfer distance (CD) and Hausdorff distance (HD)~\cite{DBLP:journals/ral/MuleroBACTK24} as evaluation metrics because it is possible to compare unordered point clouds without requiring explicit correspondences, making them efficient for evaluating discrepancies between real and simulated fabrics. CD measures the average closest-point discrepancy, making it more tolerant to small variations, while HD captures the maximum error, emphasizing worst-case deviations. 

%% file: sections/results.tex
\section{Results \& Discussion}

\subsection{DIFFCLOUD}\label{sec:Diffcloud}

DiffSim (simulator for DIFFCLOUD) models the fabric as a 2D mesh with 81 vertices, with handle vertex positions keyframed to mimic real-world motion over 60 timesteps. For \textit{wind blowing} scenarios, optimization is based on the average one-sided Chamfer distance computed over all frames. For \textit{lifting} scenarios, we compute the loss using only the last 30 frames since significant fabric deformation occurs in the second half of the sequence. All training settings (optimizer, learning rate etc.) follow those specified in the original paper~\cite{sundaresan2022diffcloud}.

We evaluated DIFFCLOUD using the \textit{lift} and \textit{wind blowing} scenarios to determine which \textit{real2sim} conditions yield better accuracy and generalization across scenarios. 
The results demonstrate consistently low Chamfer distances and stable performance, with a minimal coefficient of variation in parameter estimation. Figure~\ref{fig:real2sim} illustrates how the pipeline effectively captures fabric stiffness characteristics: high initial stiffness parameters were optimized to lower values, resulting in deformation behavior that closely matches the real point cloud data. 

While DIFFCLOUD demonstrates several strengths, it also has two limitations. First, using Chamfer distance as a metric effectively captures local geometric features but fails to represent global properties, leading to good local matching between simulated and real fabric regions without optimal overall correspondence. Second, the rendering process in the differentiable pipeline relies on backpropagating point information through a fixed mesh topology, preventing the use of remeshing techniques that could adaptively refine the mesh in highly deformable regions. This limitation affects both optimization speed and the overall simulation accuracy.

\subsection{DiffCP}\label{sec:Diffcpsix}

\begin{figure}[t]
  \centering
  \includegraphics[width=0.45\textwidth]{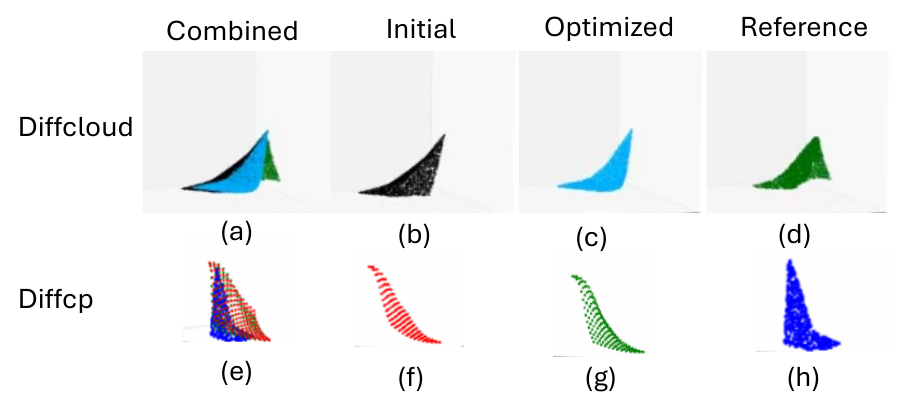}
  \caption{Visualization of DiffCloud and DiffCP approaches comparing fabric deformation states during \textit{lifting}.
  }
  \label{fig:real2sim}
\end{figure}

For DiffCP, we use the same mesh, hand vertex positions, timesteps, \textit{lifting} scenarios and training strategy as described in Sec.~\ref{sec:Diffcloud}. During training, we observed significant instability in the optimization process. The physics parameter estimation exhibited high variance across training iterations, with unstable gradients and occasional loss explosions. Moreover, the simulation struggled to accurately capture the behavior of highly deformable fabrics. As shown in Fig.\ref{fig:real2sim} and video demonstration\footnote{\url{https://cvas-ug.github.io/real2sim-study}}, although the optimized parameters produce noticeably softer fabric behavior compared to the initial settings, the simulated deformations do not fully match the deformation observed in real fabrics.

\subsection{Physnet}

We created training data as described in~\cite{duan2022learning}. DiffSim simulates \textit{wind blowing} and \textit{lifting} scenarios, and DiffTaichi, \textit{lifting} scenarios. We captured a 60 frame sequence for each scenario and fabric type. 
We use the default values in~\cite{duan2022learning} for the training process; network architecture and hyperparameters. Training results show effective differentiation of fabric physics, with fabrics forming distinct clusters. During the testing phase, we captured real fabric motion using a Zed2 camera under wind force. 

Across all experiments, the physics similarity between simulated and real observation was minimized; however, the final similarity observes high variance. This instability stems from the nature of the Bayesian optimization process as it optimizes multiple material parameters (stiffness tensors or Young's modulus and Poisson's ratio) simultaneously, focusing solely on minimizing the embedding distance between simulated and real data. As a result, the embedding distances during optimization exhibit significant fluctuations rather than steady convergence, indicating the optimizer's emphasis on numerical minimization over physical meaningfulness.

\begin{figure*}[t]
   \centering
   \includegraphics[width=0.85\linewidth]{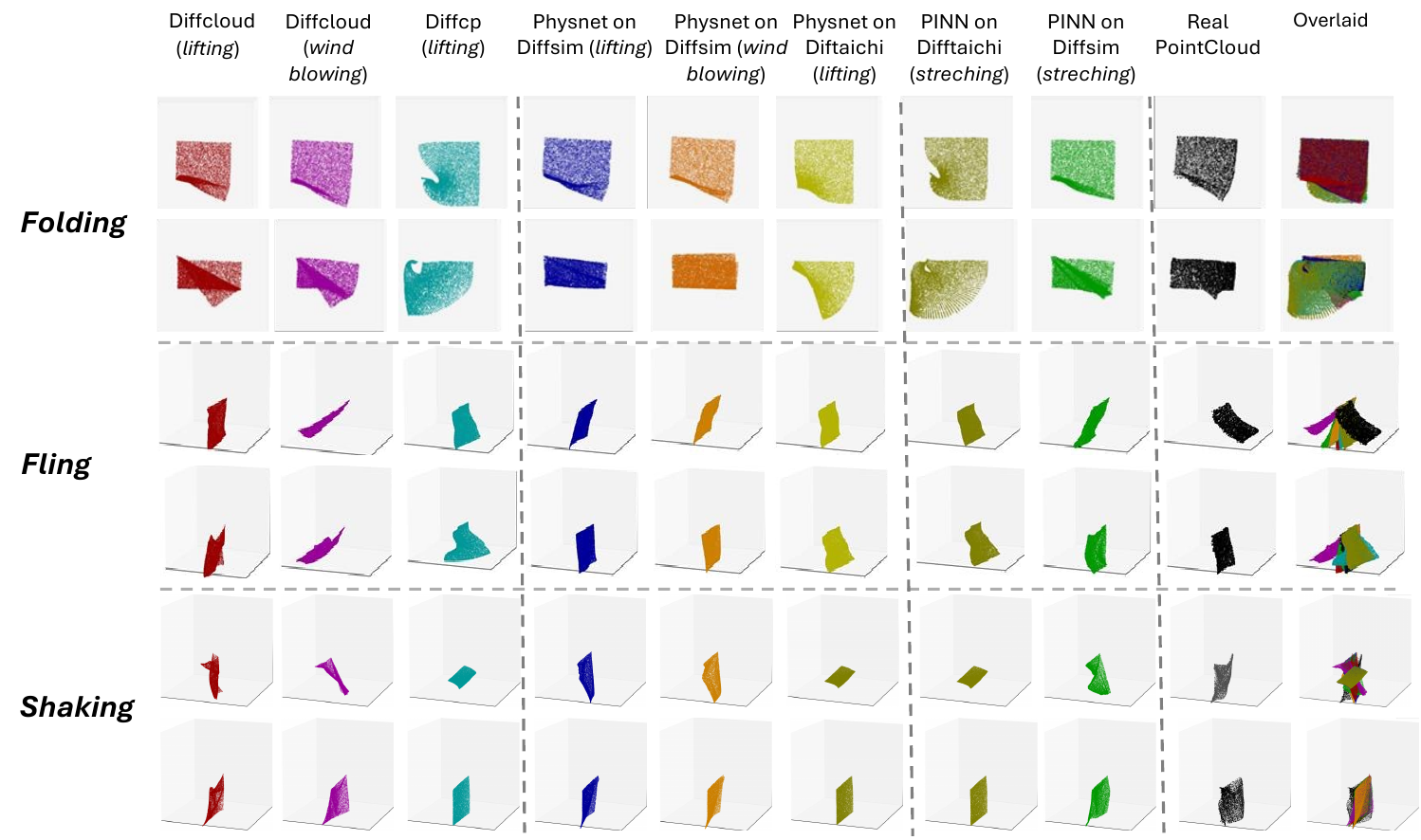}
   \caption{Comparative visualization of fabric manipulation across different Real2Sim approaches and scenarios. The three rows represent different manipulation tasks: \textit{folding} (top), \textit{fling} (middle), and \textit{shaking} (bottom). The rightmost columns show the real point cloud data and an overlaid comparison. Each approach is visualized using a distinct color for clear differentiation. The \textit{folding} sequence shows the fabric's deformation from flat to folded states, while the \textit{fling} and \textit{shaking} sequence captures the dynamic motion of the fabric during the \textit{fling} and \textit{shaking} action. 
   }\label{fig:sim2real}
\end{figure*}

\subsection{PINN--DiffSim and DiffTaichi}
The data collection procedure for PINN differs from differential pipelines and data-driven approaches because instead of relying on the geometry shape (i.e. point cloud) of the fabric, it needs corresponding points during motion. Therefore, as in Sec. \ref{sec:pinn2}, the fabric is divided into 81 sections with equidistant markers placed across its surface. The test setup involves clamping both edges with wooden strips: one edge remains fixed while the other is attached to a tensile testing device. The device extends the fabric to a predetermined displacement and maintains this position to measure steady-state force. Due to the rapid deformation process, we adjusted the camera capture frequency to 30 fps to ensure adequate temporal resolution. We collected 60 test sequences per fabric type. For point tracking, we manually annotated marker positions in each frame.

The model architecture consists of a seven-layer multi-layer perceptron (MLP)., 
We implement separate optimization strategies for data and PDE losses: Adam optimizer with a learning rate of $1 \times 10^{-4}$ for data loss, and OFBGS optimizer~\cite{DBLP:journals/corr/abs-2006-08472} with a learning rate of 0.1 for PDE loss, where stiffness parameters are learned during optimization. The training process utilizes an annealing schedule for loss weights, initializing data loss weight at 1.0 and PDE loss weight at 0.1. The PDE loss weight incrementally increases by 0.1 every 500 epochs until reaching 1.0, as in practice, we found better convergence when prioritizing physics loss reduction early in training. Each fabric is trained for 10,000 epochs.

\begin{table}[t]
\caption{Quantitative results for \textit{Folding} scenario evaluation}
\label{tab:sim-to-real1}
\centering
\resizebox{0.48\textwidth}{!}{
\begin{tabular}{@{}clcccccccc@{}}
\toprule 
    & & \multicolumn{2}{c}{\textbf{DIFFCLOUD}}&  \textbf{DiffCP} &\multicolumn{3}{c}{\textbf{PhysNet}} & \multicolumn{2}{c}{\textbf{PINN}}  \\
\cmidrule(r){3-4}\cmidrule(r){5-5}\cmidrule(r){6-8}\cmidrule{9-10}
& & L& W& L& DS,L& DS,W &DT,L &DT,S &DS,S\\
\midrule
\multirow{6}{*}{\textbf{CD}}
    & Camel Ponte Roma & 0.493 & 0.028 & 0.118 & 0.039 & 0.026 & 0.337 & 0.452 & \textbf{0.017} \\
    & Black Denim & \textbf{0.018} & 0.042 & 0.115 & 0.018 & 0.022 & 0.283 & 0.192 & 0.028 \\
    & Grey Interlock & 0.011 & 0.039 & 0.075 & 0.027 & 0.036 & 0.108 & 0.093 & \textbf{0.009} \\
    & Pink Solid & 0.040 & 0.018 & 0.101 & 0.047 & 0.050 & 0.258 & 0.143 & \textbf{0.010} \\
    & Red Jet Set & 0.013 & 0.023 & 0.170 & 0.014 & 0.016 & 0.471 & 0.299 & \textbf{0.012} \\
    & \textbf{Average} & 0.115 & 0.030 & 0.116 & 0.029 & 0.030 & 0.291 & 0.234 & \textbf{0.015} \\
\hline
\rule{0pt}{8pt}
\multirow{6}{*}{\textbf{HD}} 
    & Camel Ponte Roma & 0.779 & 0.582 & 1.428 & 0.836 & 0.689 & 1.501 & 1.606 & \textbf{0.265} \\
    & Black Denim & 0.270 & 0.792 & 1.255 & 0.265 & \textbf{0.247} & 1.723 & 1.556 & 0.739 \\
    & Grey Interlock & 0.312 & 0.783 & 1.144 & 0.623 & 0.781 & 1.249 & 0.984 & \textbf{0.309} \\
    & Pink Solid & 0.938 & 0.410 & 1.033 & 0.932 & 0.956 & 1.311 & 1.418 & \textbf{0.286} \\
    & Red Jet Set & 0.530 & 0.419 & 0.947 & 0.432 & \textbf{0.384} & 1.781 & 1.520 & 0.406 \\
    & \textbf{Average} & 0.566 & 0.597 & 1.161 & 0.618 & 0.611 & 1.513 & 1.417 & \textbf{0.401} \\
\bottomrule
\end{tabular}}

\vspace{2pt}
\hfill \tiny{\textit{CD: Chamfer Distances; HD: Hausdorff Distances; L: Lift; W: Wind blowing; S: Stretch; DS: DiffSim; DT: DiffTaichi}} 
\end{table}

\subsection{Evaluation}

The goal of these experiments is to determine the generalization ability of the implemented \textit{real2sim} approaches in Sec. \ref{sec:methods} while evaluating estimated physics parameters in the evaluation scenarios. 
Results and visualizations are shown in Tables~\ref{tab:sim-to-real1}, \ref{tab:sim-to-real2} and \ref{tab:sim-to-real3}, and Fig.\ref{fig:sim2real}, respectively.

\subsubsection{Real2Sim simulation's influence}

Tables~\ref{tab:sim-to-real1}, ~\ref{tab:sim-to-real2} and~\ref{tab:sim-to-real3} reveal significant limitations for all the fabric types utilizing the DiffTaichi simulator due to the isotropic material model representation, which enforces uniform physical behavior across all directions. This simplified representation is inadequate for capturing complex non-linear deformations characteristic of real fabrics. For example, from Tab. \ref{tab:sim-to-real1}, during \textit{folding}, the simulated fabric exhibits excessive rigidity, and the Chamfer Distance and Hausdorff Distance metrics are high compared to other approaches. While in \textit{fling} and \textit{shaking} motions, the fabric maintains deformed states reminiscent of elastic materials rather than demonstrating natural fabric recovery behavior. In contrast, the DiffSim simulator demonstrates consistent performance across various \textit{real2sim} approaches and scenarios for all types of fabrics, achieving a reasonable representation of fabric behavior.

\begin{table}[t]
\caption{Quantitative results for \textit{Fling} scenario evaluation}
\label{tab:sim-to-real2}
\centering
\resizebox{0.48\textwidth}{!}{
\begin{tabular}{@{}clcccccccc@{}}
\toprule 
    & & \multicolumn{2}{c}{\textbf{DIFFCLOUD}}&  \textbf{DiffCP} &\multicolumn{3}{c}{\textbf{PhysNet}} & \multicolumn{2}{c}{\textbf{PINN}}  \\
\cmidrule(r){3-4}\cmidrule(r){5-5}\cmidrule(r){6-8}\cmidrule{9-10}
& & L& W& L& DS,L& DS,W &DT,L &DT,S &DS,S\\
\midrule
\multirow{6}{*}{\centering \textbf{CD}}
    & Camel Ponte Roma & \textbf{0.370} & 1.018 & 0.617  & 0.918 & 1.061 & 0.753 & 0.548 & 1.623 \\
    & Black Denim & \textbf{0.656} & 0.721 & 0.925 & 0.770 & 0.981 & 0.959 & 0.858 & 0.825 \\
    & Grey Interlock & 0.729 & 1.153 & 0.465 & 0.761 & 0.763 & 0.516 & \textbf{0.379} & 0.747 \\
    & Pink Solid & 0.924 & 1.828 & 0.324 & 0.863 & 0.856 & 0.783 & \textbf{0.319} & 0.964 \\
    & Red Jet Set & 0.668 & 1.446 & 0.472 & 0.712 & 0.707 & 0.951 & \textbf{0.388} & 0.956 \\
    & \textbf{Average} & 0.669 & 1.233 & 0.561 & 0.805 & 0.874 & 0.792 & \textbf{0.498} & 1.023 \\
\hline
\rule{0pt}{8pt}
\multirow{6}{*}{\textbf{HD}} 
    & Camel Ponte Roma & \textbf{1.080} & 1.511 & 1.341 & 1.421 & 1.484 & 1.432 & 1.085 & 1.950 \\
    & Black Denim & 1.241 & 1.231 & 1.258 & 1.305 & 1.454 & 1.899 & \textbf{1.211} & 1.385 \\
    & Grey Interlock & 1.198 & 1.603 & 1.236 & 1.240 & 1.236 & \textbf{0.978} & 1.016 & 1.234 \\
    & Pink Solid & 1.409 & 1.869 & 0.987 & 1.349 & 1.363 & 1.922 & \textbf{0.885} & 1.471 \\
    & Red Jet Set & 1.166 & 1.637 & 1.407 & 1.120 & 1.133 & 1.369 & \textbf{1.010} & 1.327 \\
    & \textbf{Average} & 1.219 & 1.570 & 1.246 & 1.287 & 1.334 & 1.520 & \textbf{1.041} & 1.473 \\
\bottomrule
\end{tabular}}

\vspace{2pt}
\hfill \tiny{\textit{CD: Chamfer Distances; HD: Hausdorff Distances; L: Lift; W: Wind blowing; S: Stretch; DS: DiffSim; DT: DiffTaichi}} 
\end{table}

\begin{table}[t]
\caption{Quantitative results for Shaking scenario evaluation}
\label{tab:sim-to-real3}
\centering
\resizebox{0.48\textwidth}{!}{
\begin{tabular}{@{}clcccccccc@{}}
\toprule 
    & & \multicolumn{2}{c}{\textbf{DIFFCLOUD}}&  \textbf{DiffCP} &\multicolumn{3}{c}{\textbf{PhysNet}} & \multicolumn{2}{c}{\textbf{PINN}}  \\
\cmidrule(r){3-4}\cmidrule(r){5-5}\cmidrule(r){6-8}\cmidrule{9-10}
& & L& W& L& DS,L& DS,W &DT,L &DT,S &DS,S\\
\midrule
\multirow{6}{*}{\centering \textbf{CD}}
    & Camel Ponte Roma & \textbf{0.129}  & 0.561 & 1.292 & 0.232 & 0.433 & 1.235 & 1.292 & 0.530 \\
    & Black Denim & \textbf{0.229} & 0.594 & 1.464 & 0.234 & 0.394 & 1.445 & 1.449 & 0.401 \\
    & Grey Interlock & \textbf{0.173} & 0.443 & 1.644 & 0.215 & 0.174 & 1.681 & 1.720 & 0.200 \\
    & Pink Solid & 0.310 & 0.698 & 1.032 & 0.196 & \textbf{0.192} & 1.153 & 1.204 & 0.360 \\
    & Red Jet Set & 0.233 & 0.475 & 1.445 & 0.092 & \textbf{0.091} & 1.615 & 1.485 & 0.309 \\
    & \textbf{Average} & 0.215 & 0.554 & 1.375 & \textbf{0.194} & 0.257 & 1.426 & 1.430 & 0.360 \\
\hline
\rule{0pt}{8pt}
\multirow{6}{*}{\textbf{HD}} 
    & Camel Ponte Roma & \textbf{0.786} & 1.218 & 2.002 & 0.871 & 1.079 & 1.935 & 1.997 & 1.378 \\
    & Black Denim & \textbf{1.128} & 1.572 & 2.438 & 1.165 & 1.334 & 2.427 & 2.427 & 1.363 \\
    & Grey Interlock & 0.821 & 1.210 & 2.223 & 0.881 & \textbf{0.805} & 2.265 & 2.272 & 0.878 \\
    & Pink Solid & 1.079 & 1.420 & 1.781 & 0.892 & \textbf{0.877} & 1.870 & 1.879 &0.891 \\
    & Red Jet Set & 0.936 & 1.211 & 2.066 & \textbf{0.708} & 0.710 & 2.049 & 2.097 & 1.037 \\
    & \textbf{Average} &0.950 & 1.326 & 2.102 & \textbf{0.903} & 0.961 & 2.109 & 2.134 & 1.109 \\
\bottomrule
\end{tabular}}

\vspace{2pt}
\hfill \tiny{\textit{CD: Chamfer Distances; HD: Hausdorff Distances; L: Lift; W: Wind blowing; S: Stretch; DS: DiffSim; DT: DiffTaichi}} 
\end{table}

\subsubsection{Real2Sim scenario influence}

We investigated two \textit{real2sim} scenarios (\textit{wind blowing} and \textit{lifting}) and hypothesized that the choice of the scenarios influences the accuracy of the estimation of physical behavior and generalization performance in different scenarios. To evaluate this, we analyze DIFFCLOUD and PhysNet applied to both \textit{wind blowing} and \textit{lifting} scenarios for training. The results indicate that for PhysNet, both \textit{real2sim} scenarios yield comparable performance in \textit{sim2real} transfer across all evaluation scenarios, suggesting minimal dependency on the \textit{real2sim} training scenario for fabric behavior estimation.

In contrast, DIFFCLOUD's performance varies based on material properties. Specifically, in the \textit{folding} scenario from Tab.~\ref{tab:sim-to-real1}, the wind-based scenario achieves better results for the Pink Solid fabric (CD: 0.018, HD: 0.410), whereas the \textit{lifting}-based scenario is more effective for Black Denim (CD: 0.018, HD: 0.042). Additionally, for the \textit{fling} and \textit{shaking} scenario, the \textit{lifting}-based scenario outperforms the wind-based counterpart. A possible explanation for these results is that fabric \textit{lifting} applies well-defined and concentrated tensile forces, which directly reveal fabric stiffness properties through observable deformations. In contrast, wind-induced motion generates dispersed and fluctuating forces that rarely produce consistent tensile states, making it less effective for characterizing fabric stiffness parameters. Consequently, the \textit{lifting} scenario appears to provide more reliable conditions for estimating and calibrating fabric stiffness. Therefore, the choice of \textit{real2sim} scenarios has a limited impact on the generalizability of PhysNet. This is evident from the fact that the same \textit{real2sim} approaches produce nearly identical outcomes regardless of the training scenario used. However, for DIFFCLOUD, the choice of the \textit{real2sim} scenario affects \textit{sim2real} generalization performance, particularly in the \textit{shaking} and \textit{fling} scenarios.

\subsubsection{Sim2real approach}

Our analysis across differential pipelines, PhysNet, and PINNs reveals that the \textit{real2sim} approaches selection significantly impacts \textit{sim2real} performance. For \textit{folding} scenarios (Tab. \ref{tab:sim-to-real1}), the PINN achieves the lowest Chamfer and Hausdorff distances in four out of five fabric types when using DiffSim, followed by the differential pipeline approaches, with PhysNet exhibiting the worst performance.
The \textit{fling} scenario results demonstrate dependency on both simulations and \textit{real2sim} approach combinations. The DIFFCLOUD approach achieves optimal metrics for Black Denim and Grey Interlock fabrics when using \textit{lifting} scenarios with the DiffSim simulator. PINNs performs best for the remaining fabric types when implementing stretching scenarios with the DiffTaichi simulator.

Overall, our proposed Physics-Informed Neural Network (PINN) achieves superior performance in \textit{folding} tasks, outperforming the second-best approach by a Chamfer distance margin of 0.0148 and a Hausdorff distance margin of 0.2104 as shown in Tab \ref{tab:sim-to-real1}, with its limitation in \textit{fling} and \textit{shaking} scenarios likely due to the use of stretching as the \textit{real2sim} scenario, which may inadequately capture dynamic manipulation behavior. Future work could address these limitations through boundary condition refinement ~\cite{DBLP:journals/corr/abs-2310-02548}, residual region point resampling ~\cite{DBLP:journals/jcphy/ZhangLL25}, and extension to 3D scenarios ~\cite{heger2024investigation} for PINN-based fabric parameter estimation.

%% file: sections/conclusions.tex
\section{CONCLUSIONS}

In this paper, we presented a rigorous evaluation of the generalization capabilities of various \textit{real2sim} parameter estimation approaches for fabric manipulation. Our study systematically evaluates four state-of-the-art approaches: two differential pipelines (DIFFCLOUD~\cite{sundaresan2022diffcloud} and DiffCP~\cite{{diffcp2019}}), data-driven (PhysNet), and a physics-informed neural network approach on two simulations (DiffSim~\cite{DBLP:conf/icml/QiaoLKL20} and DiffTaichi~\cite{DBLP:conf/iclr/HuALSCRD20}). These approaches are used to estimate the physics parameters of five fabric types across multiple \textit{real2sim} scenarios (\textit{lifting}, \textit{wind blowing}, and \text{stretching}). Estimated physics parameters are then used to simulate a fabric on evaluation tasks (\textit{folding}, \textit{shaking}, and \textit{fling}).

In this paper, we found that the simulation framework and constitutive material modeling play a critical role in \textit{real2sim} and evaluation performance. Similarly, the choice of \textit{real2sim} scenarios showed a comparable impact on the generalization ability of  \textit{real2sim} approaches. Finally, the selection of \textit{real2sim} approaches significantly influences fabric manipulation performance, highlighting the critical role of accurate physics parameter estimation in reducing the Sim-to-Real gap. Among the evaluated approaches, PINN demonstrated superior performance in quasi-static tasks such as folding, achieving an average error reduction of 16.5\% compared to DIFFCLOUD, the second best approach (see Table \ref{tab:sim-to-real1}). It also showed consistent results across all fabric types. However, its effectiveness diminished in dynamic scenarios, primarily due to its reliance on stretching-based parameter estimation. This study revealed three fundamental challenges: (1) differential approaches face a trade-off between local and global geometric accuracy, (2) optimization processes exhibit inherent instabilities, and (3) PINNs show the limitation on \textit{real2sim} scenario setting. We envision that these insights provide valuable guidance for future research in robotic deformable object manipulation and highlight the importance of choosing appropriate simulation environments and \textit{real2sim} methods based on specific task requirements for robotic fabric and garment manipulation.

%% file: root.bbl
\begin{thebibliography}{10}
\providecommand{\url}[1]{#1}
\csname url@samestyle\endcsname
\providecommand{\newblock}{\relax}
\providecommand{\bibinfo}[2]{#2}
\providecommand{\BIBentrySTDinterwordspacing}{\spaceskip=0pt\relax}
\providecommand{\BIBentryALTinterwordstretchfactor}{4}
\providecommand{\BIBentryALTinterwordspacing}{\spaceskip=\fontdimen2\font plus
\BIBentryALTinterwordstretchfactor\fontdimen3\font minus \fontdimen4\font\relax}
\providecommand{\BIBforeignlanguage}[2]{{%
\expandafter\ifx\csname l@#1\endcsname\relax
\typeout{** WARNING: IEEEtran.bst: No hyphenation pattern has been}%
\typeout{** loaded for the language `#1'. Using the pattern for}%
\typeout{** the default language instead.}%
\else
\language=\csname l@#1\endcsname
\fi
#2}}
\providecommand{\BIBdecl}{\relax}
\BIBdecl

\bibitem{duan2022continuous}
L.~Duan and G.~Aragon-Camarasa, ``A continuous robot vision approach for predicting shapes and visually perceived weights of garments,'' \emph{IEEE Robotics and Automation Letters}, vol.~7, no.~3, pp. 7950--7957, 2022.

\bibitem{lin2021softgym}
X.~Lin, Y.~Wang, J.~Olkin, and D.~Held, ``Softgym: Benchmarking deep reinforcement learning for deformable object manipulation,'' in \emph{Conference on Robot Learning}.\hskip 1em plus 0.5em minus 0.4em\relax PMLR, 2021, pp. 432--448.

\bibitem{elguea2023review}
{\'I}.~Elguea-Aguinaco, A.~Serrano-Mu{\~n}oz, D.~Chrysostomou, I.~Inziarte-Hidalgo, S.~B{\o}gh, and N.~Arana-Arexolaleiba, ``A review on reinforcement learning for contact-rich robotic manipulation tasks,'' \emph{Robotics and Computer-Integrated Manufacturing}, vol.~81, p. 102517, 2023.

\bibitem{matas2018sim}
J.~Matas, S.~James, and A.~J. Davison, ``Sim-to-real reinforcement learning for deformable object manipulation,'' in \emph{Conference on Robot Learning}.\hskip 1em plus 0.5em minus 0.4em\relax PMLR, 2018, pp. 734--743.

\bibitem{sundaresan2022diffcloud}
P.~Sundaresan, R.~Antonova, and J.~Bohgl, ``Diffcloud: Real-to-sim from point clouds with differentiable simulation and rendering of deformable objects,'' in \emph{2022 IEEE/RSJ International Conference on Intelligent Robots and Systems (IROS)}.\hskip 1em plus 0.5em minus 0.4em\relax IEEE, 2022.

\bibitem{du2021diffpd}
T.~Du \emph{et~al.}, ``Diffpd: Differentiable projective dynamics,'' \emph{ACM Transactions on Graphics (TOG)}, vol.~41, no.~2, pp. 1--21, 2021.

\bibitem{DBLP:conf/cvpr/RuniaGSS20}
T.~F.~H. Runia, K.~Gavrilyuk \emph{et~al.}, ``Cloth in the wind: {A} case study of physical measurement through simulation,'' in \emph{2020 {IEEE/CVF} Conference on Computer Vision and Pattern Recognition, {CVPR}}, 2020.

\bibitem{duan2022learning}
L.~Duan, L.~Boyd, and G.~Aragon-Camarasa, ``Learning physics property parameters of fabrics and garments with a physics similarity neural network,'' \emph{IEEE Access}, vol.~10, pp. 114\,725--114\,734, 2022.

\bibitem{bouman2013estimating}
K.~L. Bouman, B.~Xiao, P.~Battaglia, and W.~T. Freeman, ``Estimating the material properties of fabric from video,'' in \emph{Proceedings of the IEEE international conference on computer vision}, 2013.

\bibitem{DBLP:journals/jscic/CuomoCGRRP22}
S.~Cuomo, V.~S.~D. Cola \emph{et~al.}, ``Scientific machine learning through physics-informed neural networks: Where we are and what's next,'' \emph{J. Sci. Comput.}, vol.~92, no.~3, p.~88, 2022.

\bibitem{zhou2023transfer}
M.~Zhou and G.~Mei, ``Transfer learning-based coupling of smoothed finite element method and physics-informed neural network for solving elastoplastic inverse problems,'' \emph{Mathematics}, vol.~11, no.~11, 2023.

\bibitem{DBLP:conf/icml/QiaoLKL20}
Y.~Qiao, J.~Liang, V.~Koltun, and M.~C. Lin, ``Scalable differentiable physics for learning and control,'' in \emph{Proceedings of the 37th International Conference on Machine Learning, {ICML}}.\hskip 1em plus 0.5em minus 0.4em\relax {PMLR}, 2020.

\bibitem{DBLP:conf/iclr/HuALSCRD20}
Y.~Hu, L.~Anderson \emph{et~al.}, ``Difftaichi: Differentiable programming for physical simulation,'' in \emph{8th International Conference on Learning Representations, {ICLR}}, 2020.

\bibitem{hu2019chainqueen}
Y.~Hu \emph{et~al.}, ``Chainqueen: A real-time differentiable physical simulator for soft robotics,'' in \emph{2019 International conference on robotics and automation (ICRA)}.\hskip 1em plus 0.5em minus 0.4em\relax IEEE, 2019, pp. 6265--6271.

\bibitem{arnavaz2023differentiable}
K.~Arnavaz, M.~K. Nielsen, P.~Kry, M.~Macklin, and K.~Erleben, ``Differentiable depth for real2sim calibration of soft body simulations,'' in \emph{Computer Graphics Forum}, vol.~42, no.~1, 2023, pp. 277--289.

\bibitem{DBLP:conf/iclr/MurthyMGVPWCPXE21}
J.~K. Murthy, M.~Macklin \emph{et~al.}, ``gradsim: Differentiable simulation for system identification and visuomotor control,'' in \emph{9th International Conference on Learning Representations, {ICLR} 2021}, 2021.

\bibitem{li2022diffcloth}
Y.~Li, T.~Du, K.~Wu, J.~Xu, and W.~Matusik, ``Diffcloth: Differentiable cloth simulation with dry frictional contact,'' \emph{ACM Transactions on Graphics (TOG)}, vol.~42, no.~1, pp. 1--20, 2022.

\bibitem{DBLP:journals/corr/abs-2007-08501}
N.~Ravi, J.~Reizenstein \emph{et~al.}, ``Accelerating 3d deep learning with pytorch3d,'' \emph{CoRR}, vol. abs/2007.08501, 2020.

\bibitem{diffcp2019}
A.~Agrawal, S.~Barratt, S.~Boyd, E.~Busseti, and W.~Moursi, ``Differentiating through a cone program,'' \emph{Journal of Applied and Numerical Optimization}, vol.~1, no.~2, pp. 107--115, 2019.

\bibitem{DBLP:journals/corr/abs-2402-00326}
S.~Wang \emph{et~al.}, ``Piratenets: Physics-informed deep learning with residual adaptive networks,'' \emph{CoRR}, vol. abs/2402.00326, 2024.

\bibitem{DBLP:journals/jcphy/ZhangLL25}
Z.~Zhang, J.~Li, and B.~Liu, ``Annealed adaptive importance sampling method in pinns for solving high dimensional partial differential equations,'' \emph{J. Comput. Phys.}, vol. 521, p. 113561, 2025.

\bibitem{DBLP:journals/corr/abs-2304-03689}
A.~Farkane, M.~Ghogho, M.~Oudani, and M.~Boutayeb, ``{EPINN-NSE:} enhanced physics-informed neural networks for solving navier-stokes equations,'' \emph{CoRR}, vol. abs/2304.03689, 2023.

\bibitem{DBLP:conf/iclr/0028MNRMGM22}
J.~Xu, V.~Makoviychuk, Y.~S. Narang, F.~Ramos, W.~Matusik, A.~Garg, and M.~Macklin, ``Accelerated policy learning with parallel differentiable simulation,'' in \emph{The Tenth International Conference on Learning Representations, {ICLR} 2022}, 2022.

\bibitem{DBLP:journals/corr/abs-2006-08472}
C.~Rao \emph{et~al.}, ``Physics informed deep learning for computational elastodynamics without labeled data,'' \emph{CoRR}, vol. abs/2006.08472, 2020.

\bibitem{wang2011data}
H.~Wang, J.~F. O'Brien, and R.~Ramamoorthi, ``Data-driven elastic models for cloth: modeling and measurement,'' \emph{ACM transactions on graphics (TOG)}, vol.~30, no.~4, pp. 1--12, 2011.

\bibitem{DBLP:conf/iccv/DoerschYVG0ACZ23}
\BIBentryALTinterwordspacing
C.~Doersch, Y.~Yang, M.~Vecer{\'{\i}}k, D.~Gokay, A.~Gupta, Y.~Aytar, J.~Carreira, and A.~Zisserman, ``{TAPIR:} tracking any point with per-frame initialization and temporal refinement,'' in \emph{{IEEE/CVF} International Conference on Computer Vision, {ICCV} 2023, Paris, France, October 1-6, 2023}.\hskip 1em plus 0.5em minus 0.4em\relax {IEEE}, 2023, pp. 10\,027--10\,038. [Online]. Available: \url{https://doi.org/10.1109/ICCV51070.2023.00923}
\BIBentrySTDinterwordspacing

\bibitem{DBLP:flatnfold}
L.~Zhuang, S.~Fan, Y.~Ru, F.~P. Audonnet, P.~Henderson, and G.~Aragon{-}Camarasa, ``Flat'n'fold: {A} diverse multi-modal dataset for garment perception and manipulation,'' \emph{CoRR; accepted for publication in ICRA 2025}, vol. abs/2409.18297, 2024.

\bibitem{DBLP:journals/corr/abs-2408-00714}
N.~Ravi, V.~Gabeur \emph{et~al.}, ``{SAM} 2: Segment anything in images and videos,'' \emph{CoRR}, vol. abs/2408.00714, 2024.

\bibitem{trimble2019slip}
S.~Trimble, W.~Naeem, and S.~McLoone, ``Slip signal analysis on a baxter robot,'' in \emph{5th IFAC Conference on Intelligent Control and Automation Sciences (ICONS 2019)}, 2019.

\bibitem{armcontrolsystem}
{Rethink Robotics}, ``Arm control system,'' \url{https://support.rethinkrobotics.com/support/solutions/articles/80000980284-arm-control-system}, 2022, accessed: 2024-08-05.

\bibitem{DBLP:journals/ral/MuleroBACTK24}
D.~B. Mulero, O.~Barbany, G.~Alcan, A.~Colom{\'{e}}, C.~Torras, and V.~Kyrki, ``Benchmarking the sim-to-real gap in cloth manipulation,'' \emph{{IEEE} Robotics Autom. Lett.}, vol.~9, no.~3, pp. 2981--2988, 2024.

\bibitem{DBLP:journals/corr/abs-2310-02548}
\BIBentryALTinterwordspacing
S.~Barschkis, ``Exact and soft boundary conditions in physics-informed neural networks for the variable coefficient poisson equation,'' \emph{CoRR}, vol. abs/2310.02548, 2023. [Online]. Available: \url{https://doi.org/10.48550/arXiv.2310.02548}
\BIBentrySTDinterwordspacing

\bibitem{heger2024investigation}
P.~Heger, D.~Hilger, M.~Full, and N.~Hosters, ``Investigation of physics-informed deep learning for the prediction of parametric, three-dimensional flow based on boundary data,'' \emph{Computers \& Fluids}, vol. 278, p. 106302, 2024.

\end{thebibliography}
